# Symbolic Probabilistic Inference in large BN2O networks

**Bruce D'Ambrosio**
Department of Computer Science
Oregon State University
dambrosi@research.cs.orst.edu

## Abstract

A BN2O network is a two level belief net in which parent interactions are modeled using the noisy-or interaction model. In this paper we discuss application of the SPI local expression language [1] to efficient inference in large BN2O networks. In particular, we show that there is significant structure which can be exploited to improve over the Quickscore result. We further describe how symbolic techniques can provide information which can significantly reduce the computation required for computing all cause posterior marginals. Finally, we present a novel approximation technique with preliminary experimental results.

## 1 Introduction

In this paper we discuss application of Symbolic Probabilistic Inference (SPI) to the problem of computing disease posteriors in the QMR-DT BN2O network[1]. A BN2O network [4] is a two level network in which parent (disease) interactions at a child (symptom) are modeled using the noisy-or interaction model [6]. The QMR-DT network is a very large network, with over 600 diseases, 4000 findings, and 40,000 disease-finding links. Some findings have as many as 150 parents, and a case can have as many as 50 positive findings. Exact inference would, then, seem intractable. We show that inference is more tractable than in might appear, although the most difficult cases remain beyond reach of the techniques presented here. We begin with a review of the local expression language, an algebraic language used in SPI to decompose parent-child dependency. We then build up in layers the techniques we have developed to apply SPI to the QMR-DT inference task.

## 2 Local Expression Languages

In this section we review our local expression language, an extension to the standard representation for belief nets. This extended expression language is useful for compact representation of the noisy-or interaction model.

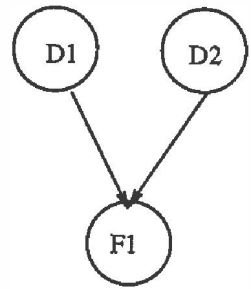

Figure 1: Noisy Or Sample Net

The local expression (that is, the expression which describes numerically the dependence of the values a variable can take on the values of its antecedents) in a belief net is simple: it is either a marginal or conditional probability distribution. While this representation is complete (that is, is capable of expressing any coherent probability model), it suffers from both space and time complexity limitations: both the space and time (for inference) required are exponential in the number of antecedents. However, computation of child marginals using the noisy-or interaction model is linear in the number of (independent) antecedents in both space and time. When evidence is available on child variables, computation of the posterior probability of parents is exponential in the number of positive pieces of evidence, but linear in the number of pieces of negative evidence, as shown in Heckerman [2]. If the interaction between the effects of $D1$ and $D2$ on $F1$ in the net shown in figure 1 can be modeled as a noisy-or interaction, then we might write the following expression for the dependence of $F1$ on $D1$ and $D2$, following Pearl [6]:

$$P(F1 = t) = 1 - (1 - c(F1|D1 = t)) \\ * (1 - c(F1|D2 = t))$$
$$P(F1 = f) = (1 - c(F1|D1 = t)) \\ * (1 - c(F1|D2 = t))$$

---

[1]This work was supported by Grant IRI-9120330 from the National Science Foundation to the Institute for Decision Systems Research, and under NSF IRI 91-00530. It benefitted greatly from discussions with Max Henion, Greg Provan, Bob Fung, and all the other IDSR folk, as well as with Tom Dietterich

Where $c(F1|D1 = t)$ is the probability that $F1$ is true given that $D1$ is true and $D2$ is false. We use $c$ rather than $p$ to emphasize that these are not standard conditional probabilities. We will use a slightly more compact notation. We can define $c'(F1|D1)$:

$$c'(F1|D1) = 1 - c(F1|D1 = t), D1 = t \\ = 1, D1 = f$$

Now we can reexpress the above as:

$$P(F1 = t) = 1 - c'(F1|D1) * c'(F1|D2)$$
$$P(F1 = f) = c'(F1|D1) * c'(F1|D2)$$

This notation is intuitively appealing. It is compact (linear in the number of antecedents), captures the structure of the interaction, and, as Heckerman has shown [2], can be manually manipulated to perform efficient inference. However, it is not sufficiently formal to permit automated inference. We define in [1] a formal syntax for a local expression language. This language permits description of the dependence of a node on its parents in terms of simple arithmetic expressions over partial distributions. Each distribution, in turn, is defined over some rectangular subspace of the cartesian product of domains of its conditioned and conditioning variables. Examining the simple noisy-or example provided earlier, we discover that the informal representation obscured the fact that the two instances of $c'(F1|D1)$ are in fact operating over disjoint domains. In the remainder of this paper we will use the following compact notation to specify expressions and distributions:[2]

[2] We ignore the actual numeric values in the distribution, since they are not germane to the discussion.

$$exp(F1) = 1_{F1_t} - c'_{F1_t|D1_{t,f}} * c'_{F1_t|D2_{t,f}} \\ + c'_{F1_f|D1_{t,f}} * c'_{F1_f|D2_{t,f}}$$

Note that in this representation there are two instances of $c_{D1}(F1)$. While the numeric distributions are identical, the domains over which they are defined differ.

We have specified a syntax and shown that a noisy-or can be expressed in this syntax. We will next review the semantics for the language and whether or not these semantics match those standardly attributed to the noisy-or structural model. Expression semantics are quite simple to specify:

> An expression is equivalent to the distribution obtained by evaluating it using the standard rules of arithmetic for each possible combination of antecedent values.

Performing this evaluation symbolically for our simple example yields:

| D1 | D2 | F=t |
|----|----|-----|
| t  | t  | $1 - (1 - c_{F1}(D1)) * (1 - c_{F1}(D2))$ |
| t  | f  | $1 - (1 - c_{F1}(D1)) * (1))$ |
| f  | t  | $1 - (1) * (1 - c_{F1}(D2))$ |
| f  | f  | $1 - (1) * (1)$ |
|    |    | F=f |
| t  | t  | $(1 - c_{F1}(D1)) * (1 - c_{F1}(D1))$ |
| t  | f  | $(1 - c_{F1}(D1)) * (1))$ |
| f  | t  | $(1) * (1 - c_{F1}(D2))$ |
| f  | f  | $(1) * (1)$ |

This is, in fact, exactly the standard semantics attributed to noisy-or.

## 3 Inference Basics

### 3.1 Evidence

SPI rewrites expressions for finding variables and immediate successors eliminating all references to unobserved values of the finding variable. As a result, the expression for a positive finding is reduced to:

$$exp(F1) = 1_{F1_t} - c'_{D2_t|D1_{t,f}} * c'_{F1_t|D2_{t,f}}$$

And that for a negative finding to:

$$exp(F1) = c'_{F1_f|D1_{t,f}} * c'_{F1_f|D2_{t,f}}$$

### 3.2 Inference

Given that, a bit of thought will reveal that negative findings can be processed in time linear in both the number of findings and the number of diseases. The remainder of the paper will concentrate on positive findings. The first stage in efficient exact inference is to note that one can distribute the disease priors over the expressions for the positive findings. Doing this, followed by application of commutativity, changes evaluation complexity from exponential in the number of diseases to exponential in the number of positive findings and linear in the number of diseases (this is the basic Quickscore result [3], re-interpreted). Since the number of positive findings is usually much lower than the number of diseases, this is advantageous. Consider, for example, a fully connected BN2O net (ie, every disease is a parent of every finding) with three diseases and two symptoms. If there are positive findings for both symptoms, the posterior expression for a disease is:

$$P(D1) = \sum_{D2,D3}$$
$$(1_{F1_t} - c'_{F1_t|D1_{t,f}} * c'_{F1_t|D2_{t,f}} * c'_{F1_t|D3_{t,f}})$$
$$* (1_{F2_t} - c'_{F2_t|D1_{t,f}} * c'_{F2_t|D2_{t,f}} * c'_{F2_t|D3_{t,f}})$$
$$* P(D1) * P(D2) * P(D3)$$

"Normal" evaluation of this expression is exponential in the number of diseases, since the full conditional for each finding is exponential in the number of parents. However, we can distribute over the finding expressions, then apply associativity and distributivity, to obtain:

$$P(D1) =$$
$$1_{F1_t} * 1_{F2_t} * P(D1) * \sum_{D2} P(D2) * \sum_{D3} P(D3)$$
$$- 1_{F1_t} * c'_{F2_t|D1_{t,f}} * P(D1)$$
$$* (\sum_{D2} c'_{F2_t|D2_{t,f}} * P(D2))$$
$$* (\sum_{D3} c'_{F2_t|D3_{t,f}} * P(D3))$$
$$- 1_{F2_t} * c'_{F1_t|D1_{t,f}} * P(D1)$$
$$* (\sum_{D2} c'_{F1_t|D2_{t,f}} * P(D2))$$
$$* (\sum_{D3} c'_{F1_t|D3_{t,f}} * P(D3))$$
$$+ c'_{F2_t|D1_{t,f}} * c'_{F1_t|D1_{t,f}} * P(D1)$$
$$* (\sum_{D2} c'_{F2_t|D2_{t,f}} * c'_{F1_t|D2_{t,f}} * P(D2))$$
$$* (\sum_{D3} c'_{F2_t|D3_{t,f}} * c'_{D1_t|D3_{t,f}} * P(D3))$$

Note that the number of terms in this expression is exponential in the number of positive findings, and the number of multiplications in each term is linear in both the number of diseases and positive findings.

## 4 Factoring posterior expressions

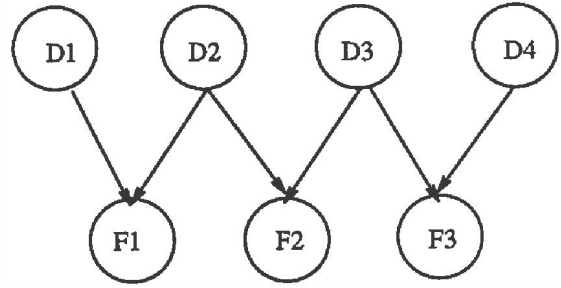

Figure 2: Sample Net for partitioning

The next step on the path to efficient exact inference in large BN2O networks is to notice that, when the network is not fully connected, distribution over the right set of findings will often enable factoring of the resulting subexpressions into independent components. We need to find a set of positive findings which form a cutset of the subgraph consisting of the positive findings and their parents. Consider a net with three findings and four diseases, as shown in fig 4. Assume we have positive findings for all three symptoms. Then the expression for the posterior distribution of $D1$ is:

$$P(D1) = \sum_{D2,D3,D4} (1_{F1_t} - c'_{F1_t|D1_{t,f}} * c'_{F1_t|D2_{t,f}})$$
$$* (1_{F2_t} - c'_{F2_t|D2_{t,f}} c'_{F2_t|D3_{t,f}})$$
$$* (1_{F3_t} - c'_{F3_t|D3_{t,f}} * c'_{F3_t|D4_{t,f}})$$
$$* P(D1) * P(D2) * P(D3) * P(D4)$$

If we distribute over F2, we get:

$$P(D1) =$$
$$\sum_{D2,D3,D4} (1_{F1_t} - c'_{F1_t|D1_{t,f}} * c'_{F1_t|D2_{t,f}})$$
$$* 1_{F2_t} * (1_{F3_t} - c'_{F3_t|D3_{t,f}} * c'_{F3_t|D4_{t,f}})$$
$$* P(D1) * P(D2) * P(D3) * P(D4)$$

$$-(1_{F1_t} - c'_{F1_t|D1_{t,f}} * c'_{F1_t|D2_{t,f}})$$
$$* c'_{F2_t|D2_{t,f}} * c'_{F2_t|D3_{t,f}}$$
$$* (1_{F1_t} - c'_{F3_t|D3_{t,f}} * c'_{F3_t|D4_{t,f}})$$
$$* P(D1) * P(D2) * P(D3) * P(D4)$$

We can now re-arrange the expression to obtain:

$$P(D1) =$$
$$\sum_{D2}(1_{F1_t} - c'_{F1_t|D1_{t,f}} * c'_{F1_t|D2_{t,f}})$$
$$* 1_{F2_t} * P(D1) * P(D2)$$
$$* \sum_{D3,D4}(1_{F3_t} - c'_{F3_t|D3_{t,f}} * c'_{F3_t|D4_{t,f}})$$
$$* P(D3) * P(D4)$$
$$- \sum_{D2}(1_{F1_t} - c'_{F1_t|D1_{t,f}} * c'_{F1_t|D2_{t,f}})$$
$$* c'_{F2_t|D2_{t,f}} * P(D1) * P(D2)$$
$$* \sum_{D3,D4}(1_{F1_t} - c'_{F3_t|D3_{t,f}} * c'_{F3_t|D4_{t,f}})$$
$$* c'_{F2_t|D3_{t,f}} * P(D3) * P(D4)$$

Notice that we have rendered independent (within each term) the computations for $D2$ and $D3, D4$. This independence comes from the fact that the graph is not a fully connected bipartite graph: the same sparseness that is exploited in standard belief net algorithms can be used here to reduce complexity of inference. The local expression language, by explicitly capturing independence at the value-specific level, permits exploitation of this structure.

One could pose an optimization question of the form we have previously posed for standard belief net inference [5]: identify that form in which an expression is least expensive to evaluate. We have work in progress to formulate the general problem for arbitrary local expressions. In this paper we present a simple greedy algorithm family based on intuitions derived from considering the above example. Given an expression to evaluate: (1) choose a positive finding expression to distribute over; (2) distribute over that expression; (3) partition each of the terms into independent sub-expressions where possible; (4) recursively, evaluate each subexpression; (5) combine the results.

There are two issues worth considering. First, how do we decide which expression to distribute first? Second, is the extra effort repaid in computational savings, especially considering that the results must be partitioned after each distribution.

**Choosing a finding to distribute over**  One myopic heuristic would be to choose that finding to distribute over which permitted the finest partitioning of each of the resulting terms. This, however, would be rather expensive to compute, and would provide no guidance in those cases where no single distribution action enables partitioning into two or more independent factors. We have found an effective heuristic that is very quick to evaluate: We distribute over any finding which has the highest number of parents. Why should such a heuristic work? Referring again to figure 4, we see that we want to distribute over the one finding that includes as parents diseases from the set {D1, D2} and the set {D3, D4}. In the absence of any further information (ie, assuming parents are randomly chosen from the set of diseases), a finding with more parents is more likely to have parents from both sets than a finding with fewer parents.

**Computational effort of partitioning**  Partitioning could be quite expensive, since the basic evaluation process is a full recursion, that is, must be applied recursively to both of the terms which result from distributing over a finding. However, we can split partitioning into two components, a relatively expensive partitioning of the remaining positive findings, and a less expensive distribution of disease priors to the appropriate partitions (we absorb finding strengths into disease priors as soon as a finding is distributed over, so they need not concern us here). Note that both terms resulting from distribution over a finding contain the same set of positive findings not yet distributed over. Therefore the expensive component of partitioning need only be done once each time we distribute over a finding. The total number of partitioning operations, therefore, is only linear in the number of positive findings.

**Other uses of partitioning**  We take as the task the computation of the marginals for all diseases that are parents of any positive finding. One way to accomplish this task is to repeat the basic computation for each such disease. Such an algorithm results in computation time that is the square of the number of diseases. The computations could be performed simultaneously by maintaining $n$ separate computation stacks, where $n$ is the number of diseases. However, consideration of the computation from an algebraic point of view reveals an alternative. We can add one additional computation stack which simply computes the prior probability of the evidence (that is, is marginalizes over every disease as early as possible, rather than holding one, the *target*, out). Then during recursive evaluation any disease that is not listed in the current expression can share the prior stack, rather than maintain a separate stack.

Typically, by the time we have incorporated 10 or 12 positive findings, all diseases are referenced in at least one finding. Therefore, we would expect computing all disease posteriors to require about 600 times as many multiplications as a single posterior (there are about 600 diseases in the QMR-DT Bn2O network). Experiments indicate that the stack sharing technique reduces the cost by a factor of 30. That is, it takes only about 20x as many multiplications to compute all disease posteriors as it does to compute a single posterior. In addition, since much of the algorithm is shared (eg, partitioning is independent of target variable, and need be done only once), the actual time increase is only about 5x.

### 4.1 Experimental Results

We performed a series of experiments to determine how much structure can be exploited in the QMR-DT BN2O network. We used a set of CPC cases and a set of Scientific American cases supplied by IDSR. Since some of these cases were too large to process in their entirety, we developed an incremental testing strategy: we posted all negative evidence, and then tested with one piece of positive evidence, two pieces of positive evidence, and so on until a time limit was exceeded. Since we were interested in exploiting structure, we ordered positive findings by number of parents, and processed first those findings with the fewest parents.

Figure 3 shows a typical trace of the partitioning which occurred in processing a case. In this figure we trace the critical portion of processing one of the cases. The trace starts when there are 17 positive findings which have not yet been distributed over. The 17 findings are not separable into independent partitions. One finding is chosen to distribute over, and the remaining sixteen are partitioned. Again, the result is a single partition. This process repeats until thirteen findings, at which point one of the findings can be partitioned from the rest (ie, has as parents only diseases not parents for any other finding). Splitting off a partition with one finding is nice (it reduces complexity by a factor of two!), but not overwhelming. The key distribution step for this case occurs two steps later, on the second distribution for the larger of the two partitions from step 13. In partitioning this set (now down to ten positive findings not yet distributed over), we find three partitions, the largest of which has only 5 findings. This step reduces evaluation complexity, then by $2^5$.

Tables 4 and 5 shows the total amount of partitioning for each case we tried. We define the total amount of partitioning to be $\sum_f(|F| - max_p|p|)$, where $F$ is the set of findings being partitioned and $p$ is the set of resulting partitions. In our example, then, the savings due to partitioning is 6. The *Done* column records whether or not we were able to process all positive findings within a 20 minute cutoff time, using a prototype common-lisp implementation running on a Sparc 2. In table 4 the column *findings* lists the total number of positive findings for cases we were able to completely process, or the number of positive findings we were able to process in 20 minutes for cases we were unable to completely process.

| Findings in term | # partitions | sizes |
|---|---|---|
| 17 | 1 | 17 |
| 16 | 1 | 16 |
| 15 | 1 | 15 |
| 14 | 1 | 14 |
| 13 | 2 | 1, 12 |
| 11 | 1 | 11 |
| 10 | 3 | 5, 4, 1 |

Figure 3: Sample Partition Sequence

| | CPC | | |
|---|---|---|---|
| Case | Pos findings | Saving | Done? |
| 1 | 29 | 11 | N |
| 2 | 24 | 5 | N |
| 3 | 20 | 1 | N |
| 4 | 23 | 4 | N |
| 5 | 23 | 5 | N |
| 6 | 22 | 4 | N |
| 7 | 22 | 3 | N |
| 8 | 24 | 7 | Y |
| 9 | 23 | 3 | N |
| 10 | 19 | 0 | N |

Figure 4: Partitioning Statistics - CPC

| | Sci-Am | | |
|---|---|---|---|
| Case | Pos findings | Saving | Done? |
| 1 | 19 | 0 | N |
| 2 | 9 | 1 | Y |
| 3 | 17 | 3 | Y |
| 4 | 8 | 0 | Y |
| 5 | 14 | 0 | Y |
| 6 | 10 | 0 | Y |
| 7 | 7 | 1 | Y |
| 8 | 8 | 0 | Y |
| 9 | 20 | 3 | N |
| 10 | 16 | 8 | Y |

Figure 5: Partitioning Statistics - Sci Am

Figure 6 show the actual execution statistics for the 10 CPC cases. This table provides some evidence that the computational gains from the methods described are real. Heckerman reported in [2] processing 9 positive findings in one minute. In 32 minutes, then quickscore should be able to process about 14 positive findings. Comparing run times of different implementations on different platforms is a difficult task, and conclusions must be carefully drawn. Nonetheless, the table provides some evidence that the benefits of the methods described here far outweigh the computational overhead involved. Execution time (in CPU minutes) and number of multiplications (in millions of floating point multiplies) are closely correlated. The number of floating point operations is about twice this number (there is roughly one addition or subtraction for each multiplication. Assuming that a Sparc 2 is roughly a 2Mflop machine, the algorithm is delivering about $\frac{1}{30}$ the raw floating point performance available. A comparable Quickscore implementation in common lisp delivered slightly less than twice this.

These experiments show that the amount of structure we are able to exploit varies widely, from a high of 11 to a low of zero. The Scientific American cases seem particularly difficult. While they tend to have fewer positive findings, those findings tend to be more complex, that is, involve more parents.

| Case | Pos findings | Done? | CPU mins | Mult |
|---|---|---|---|---|
| 1 | 29 | N | 45 | 149 |
| 2 | 24 | N | 46 | 172 |
| 3 | 20 | N | 35 | 135 |
| 4 | 23 | N | 40 | 136 |
| 5 | 23 | N | 38 | 117 |
| 6 | 22 | N | 45 | 149 |
| 7 | 22 | N | 50 | 155 |
| 8 | 24 | Y | 25 | 113 |
| 9 | 23 | N | 44 | 136 |
| 10 | 19 | N | 40 | 117 |

Figure 6: CPC run times

QMR-DT is a particularly richly connected BN20 network. Even for this network symbolic techniques are able to find some structure to exploit, but not enough to render exact inference tractable in all cases.

## 5 Incremental Refinement of Posteriors

It is interesting to note [3] that those findings that are most complex to handle are also typically least diagnostic. It is the positive findings with many parents that create computational difficulties. If this is not clear, consider the extreme opposite case where each positive finding has only one parent. These findings create no difficulty. Yet it is the findings with many parent that tend to be least diagnostic, since they implicate almost every disease. Some findings have as many as 150 parents! Worse, these finding tend to show up very frequently, probably due to the fact that almost every disease can cause them.

If positive findings with many parents are both difficult to process and not very informative, why not ignore them? We performed the following experiment to evaluate the potential of this technique: we processed the first eight positive findings (ordered by number of parents, lowest first), then used a heuristic to choose the next finding to process. The heuristic balanced difficulty of processing, as indicated by number of parents, with informativeness, estimated as the inverse of the finding prior, given findings already processed. The actual heuristic used was: $prior * \frac{\sqrt{numParents}}{100}$

To evaluate the potential of the method, we established several metrics: the error in the posterior probability of the most likely disease, the number of findings processed before the disease with the highest posterior settled permanently into its correct position, the number of findings processed before the top four settled permanently into correct positions, and the number processed before the top four were all permanently in the top four (ie, perhaps not ordered correctly among themselves). All these metrics have problems; for example, if two diseases have very close posteriors, they may exchange positions frequently. Worse, since in most of the CPC cases we were unable to process all findings, we could only use as a gold standard the results for the largest number of findings we could process. Consequently, the results can only be considered suggestive and preliminary. Figure 7 shows the results for the 10 CPC cases.

The results in figure 7 seem mixed. In most cases, we were surprised at how quickly the most likely disease settled into the number one spot. The correlation between the lowest remaining prior and the error in the disease posteriors is strongly negative. There is weaker, but still significant, evidence that diseases settle into place in order. That is, the

---

[3]I'm not sure who first observed this, but I first became aware of it in a conversation with Bob Fung.

| Case | P Finds | 1 IP | Error |
|------|---------|------|-------|
| 1 | 29 | 5 | .00001 |
| 2 | 22 | 1 | 0 |
| 3 | 20 | 1 | 0 |
| 4 | 23 | 1 | .0016 |
| 5 | 23 | 1 | 0 |
| 6 | 22 | 16 | .11 |
| 7 | 22 | 1 | .04 |
| 8 | 24 | 5 | .6 |
| 9 | 23 | 20 | .028 |
| 10 | 19 | 3 | .22 |

| Case | 4 IS | 4 IP | LEP | FLEP |
|------|------|------|-----|------|
| 1 | 17 | 28 | .0012 | .4608 |
| 2 | 20 | 21 | .0008 | .1262 |
| 3 | 6 | 17 | .0006 | .6628 |
| 4 | 17 | 23 | .0006 | .3067 |
| 5 | 21 | 22 | .0004 | .3134 |
| 6 | 12 | 21 | .4027 | .8214 |
| 7 | 12 | 22 | .0014 | .3908 |
| 8 | 21 | 22 | .0032 | .5208 |
| 9 | 22 | 23 | .2183 | .2146 |
| 10 | 18 | 18 | .0028 | .2882 |

- P Finds: Total number of positive findings processed.

- 1 IP: the finding at which the disease with highest posterior settled permanently into top position.

- Error: the difference between the disease posterior when it first appears in the number 1 spot and its final posterior. In general (and in all cases where the error is > 0.04) the final posterior is higher.

- 4 IS: the finding at which the top four diseases settled into the top four spots, possible incorrectly ordered among themselves.

- 4 IP: the finding at which top four assumed correct internal order.

- LEP: the lowest unprocessed finding prior (given evidence processed so far) at 1IP.

- FLEP: the lowest unprocessed finding prior at the end of processing (in general we stopped before processing all positive findings)

Figure 7: Incremental Finding Processing

disease with the highest posterior tends to settle into first place before the disease with the second highest posterior, and so on. It seems that in most cases a single dominant early finding (ie, one with few parents) is determining the most likely disease. Cases 6 and 9 are exceptions to this behavior. We do not yet understand why they behave differently (eg, is it the cumulative effect of several later findings, or a single later finding, or the lack of a single dominant early finding, or ...). Similarly, for most cases, the value of the lowest finding prior at which the most likely disease settles into position is very low. We had hoped to use this value predictively: once the value rose above a certain threshold, we could be assured that, with high reliability, we had identified the most likely disease. This seems to be largely the case, but again, cases 6 and 9 are exceptions.

More recently we have experimented with the Kullback-Liebler divergence measure of the difference between two distributions. Figure 8 is an example of the application of this measure to the first case from the Scientific American case set. The two curves report the K-L distance between the posteriors given the number of postive findings shown on the X axis and the final posteriors (after processing 18 findings for this case). The lower curve is for the incremental processing described here, the upper curve is for processing the positive findings in the reverse of the recommended order. The difference between the two curves is clear evidence that some findings have significantly greater impact than others, and that the ordering described here does well at identifying those findings. These results are quite preliminary, we expect to have a more complete analysis at the conference.

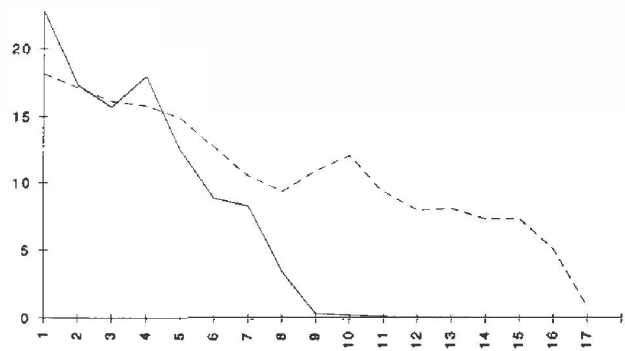

Note: solid line is normal, dashed in inverted order

Figure 8: Posterior convergence

**Discussion** This approach raises several issues. First, approximation algorithms are normally cast as methods which process all evidence approximately. This algorithm processes a subset of the evidence exactly. It depends on yet unformalized characteristics of the evidence set. We hope to have more to say about this in the final paper. Second, the task has shifted slightly. One of our metrics is a probability error metric, but most are ordering metrics. Is ordering, especially when restricted to the most likely few causes, a useful task definition? Finally, our heuristic for choosing the next finding to process is the best of a small handful we tried, but clearly can be refined further - we have not yet tried to find an optimal ordering for the findings. Such a gold standard would allow us to determine if anomalies like cases 6 and 9 are failures of the heuristic or essential to the individual cases. other classes of approximation algorithms?

## 6  Conclusion

We have shown that there is considerable structure in the QMR-DT network, and that this structure can be exploited to make inference more tractable. The results show that the QMR-DT network is at the very limit of current capability for exact computation. Networks with fewer causes, positive findings, or cause/symptom links, should be quite tractable. However, larger networks will require either changes in the basic model or approximate inference methods.